\documentclass[final]{cvpr}

\usepackage{times}
\usepackage{epsfig}
\usepackage{graphicx}
\usepackage{amsmath}
\usepackage{amssymb}
\usepackage{booktabs}
\usepackage{multirow}
\usepackage[marginal]{footmisc}

\usepackage[pagebackref=true,breaklinks=true,colorlinks,bookmarks=false]{hyperref}

\def\cvprPaperID{8484} 
\def\confYear{CVPR 2021}

\begin{document}

\title{Understanding the Behaviour of Contrastive Loss}

\makeatletter
\renewcommand*{\@fnsymbol}[1]{\ensuremath{\ifcase#1\or \dagger\or \dagger\or \ddagger\or
   \mathsection\or \mathparagraph\or \|\or **\or \dagger\dagger
   \or \ddagger\ddagger \else\@ctrerr\fi}}
\makeatother

\author{
Feng Wang, \quad Huaping Liu \thanks{Corresponding author.}\\
Beijing National Research Center for Information Science and Technology(BNRist),\\
Department of Computer Science and Technology, Tsinghua University\\
{\tt\small wang-f20@mails.tsinghua.edu.cn, hpliu@tsinghua.edu.cn}
}

\maketitle


\begin{abstract}
   Unsupervised contrastive learning has achieved outstanding success, while the mechanism of contrastive loss has been less studied. In this paper, we concentrate on the understanding of the behaviours of unsupervised contrastive loss. We will show that the contrastive loss is a hardness-aware loss function, and the temperature $\tau$ controls the strength of penalties on hard negative samples. The previous study has shown that uniformity is a key property of contrastive learning. We build relations between the uniformity and the temperature $\tau$. We will show that uniformity helps the contrastive learning to learn separable features, however excessive pursuit to the uniformity makes the contrastive loss not tolerant to semantically similar samples, which may break the underlying semantic structure and be harmful to the formation of features useful for downstream tasks. This is caused by the inherent defect of the instance discrimination objective. Specifically, instance discrimination objective tries to push all different instances apart, ignoring the underlying relations between samples. Pushing semantically consistent samples apart has no positive effect for acquiring a prior informative to general downstream tasks. A well-designed contrastive loss should have some extents of tolerance to the closeness of semantically similar samples. Therefore, we find that the contrastive loss meets a uniformity-tolerance dilemma, and a good choice of temperature can compromise these two properties properly to both learn separable features and tolerant to semantically similar samples, improving the feature qualities and the downstream performances. 
\end{abstract}

\section{Introduction}

\begin{figure}[t]
\centering
\includegraphics[width=1.0\linewidth]{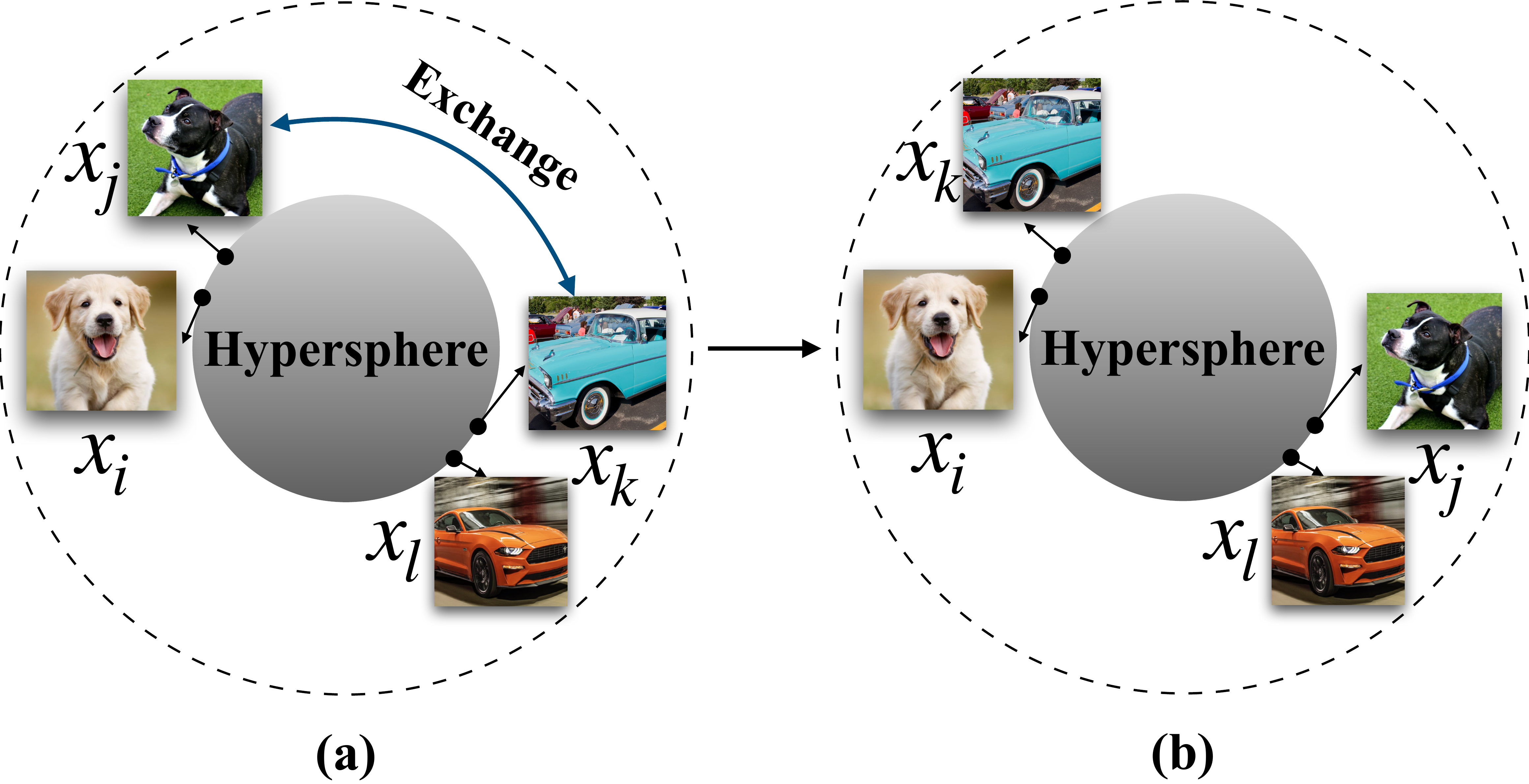}
\caption{We display two embedding distributions with four instances on a hypersphere. From the figure, we observe that exchanging $x_j$ and $x_k$, as well as their corresponding augmentations, will not change the value of contrastive loss. However, the embedding distribution of (a) is much more useful for downstream tasks because it captures the semantical relations between instances.}
\label{fig: concept}
\end{figure}

Deep neural networks have undergone dramatic progress since the large scale human-annotated datasets such as ImageNet \cite{deng2009imagenet} and Places \cite{zhou2014learning}. Such progress is heavily dependent on manual labelling, which is costly and time-consuming. Unsupervised learning gives us the promise to learn transferable representations without human supervision. Recently, unsupervised learning methods based on the contrastive loss \cite{wu2018unsupervised, oord2018representation, bachman2019amdim, he2019momentum, chen2020mocov2, chen2020simple, huang2019unsupervised, zhuang2019local} have achieved outstanding success and received increasing attention. Contrastive learning methods aim to learn a general feature function which maps the raw pixel into features residing on a hypersphere space. They try to learn representations invariant to different views of the same instance by making positive pairs attracted and negative pairs separated. With the help of heavy augmentations and strong abstraction ability of convolutional neural networks \cite{krizhevsky2012imagenet, simonyan2014very, he2016deep}, the unsupervised contrastive models can learn some extents of semantic structures. For example, in Fig \ref{fig: concept}, a good contrastive learning model tends to produce the embedding distribution likes Fig \ref{fig: concept} (a) instead of the situation of Fig \ref{fig: concept} (b), though the losses of Fig \ref{fig: concept} (a) and Fig \ref{fig: concept} (b) are the same.

Contrastive learning methods share a common design of the loss function which is a softmax function of the feature similarities with a temperature $\tau$ to help discriminate positive and negative samples. The contrastive loss is significant to the success of unsupervised contrastive learning. In this paper, we focus on analyzing the properties of the contrastive loss using the temperature as a proxy. We find that the contrastive loss is a hardness-aware loss function which automatically concentrates on optimizing the hard negative samples, giving penalties to them according to their hardness. The temperature plays a role in controlling the strength of penalties on the hard negative samples. Specifically, contrastive loss with small temperature tends to penalize much more on the hardest negative samples such that the local structure of each sample tends to be more separated, and the embedding distribution is likely to be more uniform. On the other hand, contrastive loss with large temperature is less sensitive to the hard negative samples, and the hardness-aware property disappears as the temperature approaches $+\infty$. The hardness-aware property is significant to the success of the softmax-based contrastive loss, with an explicit hard negative sampling strategy, a very simple form of contrastive loss works pretty well and achieves competitive downstream performances.

\begin{figure}[t]
\centering
\includegraphics[width=0.80\linewidth]{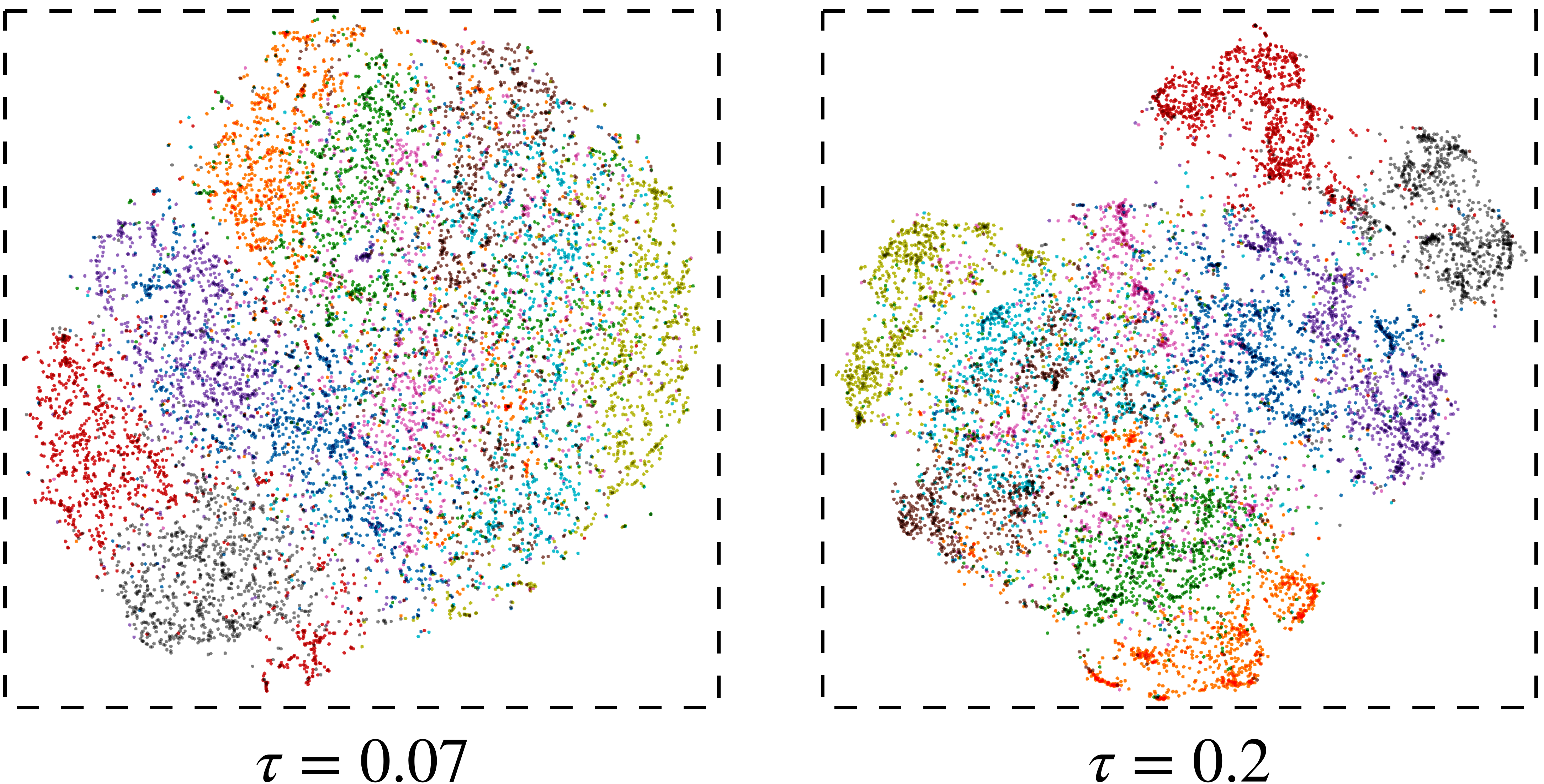}
\caption{T-SNE \cite{Maaten2008VisualizingDU} visualization of the embedding distribution. The two models are trained on CIFAR10. The temperature is set to $0.07$ and $0.2$ respectively. Small temperature tends to generate more uniform distribution and be less tolerant to similar samples.}
\label{fig: embedding}
\end{figure}

The uniformity of the embedding distribution in unsupervised contrastive learning is important to learn separable features \cite{wang2020hypersphere}. We connect the relation between the temperature and the embedding uniformity. With the temperature as a proxy, we find that although the uniformity is a key indicator to the performance of contrastive models, the excessive pursuit to the uniformity may break the underlying semantic structure. This is caused by the inherent defect of the popular unsupervised contrastive objective. Specifically, most contrastive learning methods aim to learn an instance discrimination task, by maximizing the similarities of different augmentations sampling from the same instances and minimizing the similarities of all different instances. This kind of objective actually contains no information about semantical relations. Pushing the semantically consistent samples away is harmful to generate useful features. If the contrastive loss is equipped with very small temperature, the loss function will give very large penalties to the nearest neighbours which are very likely to share similar semantical contents with the anchor point. From Fig \ref{fig: embedding}, we observe that embeddings trained with $\tau=0.07$ are more uniformly distributed, however the embeddings trained with $\tau=0.2$ present a more reasonable distribution which is locally clustered and globally separated. We recognize that there exists a uniformity-tolerance dilemma in unsupervised contrastive learning. On the one hand, we hope the features are distributed uniformly enough to be more separable. On the other hand, we hope the contrastive loss can be more tolerant to the semantically similar samples. A good contrastive loss should make a compromise to satisfy both the two properties properly. 

Overall, the contributions can be summarized as follows:
\begin{itemize}
\item We analyze the behaviours of the contrastive loss and show that contrastive loss is a hardness-aware loss. We validate that the hardness-aware property is significant to the success of contrastive loss. 

\item With a gradient analysis, we show that the temperature is a key parameter to control the strength of penalties on hard negative samples. Quantitative and qualitative experiments are conducted to validate the perspective. 

\item We show that there exists a uniformity-tolerance dilemma in contrastive learning, a good choice of temperature can compromise the two properties and improve the feature quality remarkably.
\end{itemize}

\section{Related Work}
Unsupervised learning methods have achieved great progress. Previous works focus on the design of novel pretext tasks, such as context prediction \cite{doersch2015unsupervised}, jigsaw puzzle \cite{noroozi2016unsupervised}, colorization \cite{zhang2016colorful, larsson2016learning}, rotation \cite{gidaris2018unsupervised}, context encoder \cite{pathak2016context}, split brain \cite{zhang2017split}, deep cluster \cite{Caron_2018_ECCV, caron2019unsupervised} etc. The core idea of the above self-supervised methods is to capture some common priors between the pretext task and the downstream tasks. They assume that finishing the well-designed pretext tasks requires knowledge useful for downstream tasks such as classification \cite{krizhevsky2012imagenet}, detection \cite{girshick2015fast, ren2015faster}, segmentation \cite{ronneberger2015u, he2017mask} etc. Recently, unsupervised methods based on contrastive learning have drawn increasing attentions due to the excellent performances. Wu et al \cite{wu2018unsupervised} propose an instance discrimination method, which first incorporates a contrastive loss (called NCE loss) to help discriminate different instances. CPC \cite{oord2018representation, henaff2019data} tries to learn context-invariant representations, and give a perspective of maximizing mutual information between different levels of features. CMC \cite{tian2019contrastive} is proposed to learn representations by maximizing the mutual information between different color channel views. SimCLR \cite{chen2020simple} simplifies the contrastive learning by only using different augmentations as different views, and tries to maximize the agreement between views. Besides, some methods try to maximize the agreement between different instances which may share similar semantic contents to learn instance-invariant representations, such as nearest neighbours discovery \cite{huang2019unsupervised}, local aggregation \cite{zhuang2019local}, invariance propagation \cite{wang2020unsupervised}, etc. On the other hand, contrastive loss requires many negative samples to help boost the performances. Instance discrimination \cite{wu2018unsupervised} first proposes to use a memory bank to save the calculated features as the exponential moving average of the historical features. MoCo \cite{he2019momentum, chen2020mocov2} proposes to use a momentum queue to improve the consistency of the saved features. 

There are also some works that try to understand the contrastive learning. Arora et al \cite{saunshi2019theoretical} present a theoretical framework for analyzing the contrastive learning by introducing latent classes and connect the relation between the unsupervised contrastive learning tasks and the downstream performances. Purushwalkam et al \cite{purushwalkam2020demystifying} try to demystify the unsupervised contrastive learning by focusing on the relation of data augmentation and the corresponding invariances. Tian et al \cite{tian2020makes} study the task-dependent optimal views of contrastive learning by a perspective of mutual information. Wu et al \cite{wu2020mutual} give a systematical analysis to the relations between different contrastive learning methods and the corresponding forms of mutual information. Wang et al \cite{wang2020hypersphere} try to understand the contrastive learning by two key properties, the alignment and uniformity. Different from the above works, we focus mainly on the inherent properties of the contrastive loss function. We emphasize the significance of the temperature $\tau$, and use it as a proxy to analyze some intriguing phenomenons of the contrastive learning.

\section{Hardness-aware Property}

Given an unlabeled training set $X = \{x_1, ..., x_N\}$, the contrastive loss is formulated as:
\begin{equation}
\mathcal{L}(x_i) = - {\rm log}  \left[ \frac{ {\rm exp} (s_{i,i}/\tau)}{\sum_{k \neq i} {\rm exp} (s_{i,k}/\tau) + {\rm exp} (s_{i,i}/\tau)} \right] 
\label{equation: conloss}
\end{equation}
where $s_{i,j} = f(x_i)^Tg(x_j)$. $f(\cdot)$ is a feature extractor which maps the images from pixel space to a hypersphere space. $g(\cdot)$ is a function which can be same as $f$ \cite{chen2020simple}, or comes from a memory bank \cite{wu2018unsupervised}, momentum queue \cite{he2019momentum}, etc. For convenience, we define the probability of $x_i$ being recognized as $x_j$ as:
\begin{equation}
P_{i,j} = \frac{ {\rm exp} (s_{i,j}/\tau) }{ \sum_{k \neq i} {\rm exp} (s_{i,k}/\tau) + {\rm exp}(s_{i,i}/\tau) }
\end{equation} 

The contrastive loss tries to make the positive pairs attracted and the negative samples separated, i.e., the positive alignment and negative separation. This objective can also be achieved by using a more simple contrastive loss as:
\begin{equation}
\mathcal{L}_{simple}(x_i) = -s_{i,i}+ \lambda \sum_{i \neq j}s_{i,j}
\label{eq: lsimple}
\end{equation}
However, we find that the above loss function performs much worse than the softmax-based contrastive loss of Eq \ref{equation: conloss}. In the following parts, we will show that different with $\mathcal{L}_{simple}$, the softmax-based contrastive loss is a hardness-aware loss function, which automatically concentrates on separating more informative negative samples to make the embedding distribution more uniform. Besides, we also find that the $\mathcal{L}_{simple}$ is a special case by approaching the temperature $\tau$ to $+\infty$. Next, we will start with a gradient analysis to explain the properties of the contrastive loss. 

\begin{figure}
    \centering
    \includegraphics[width=1.0\linewidth]{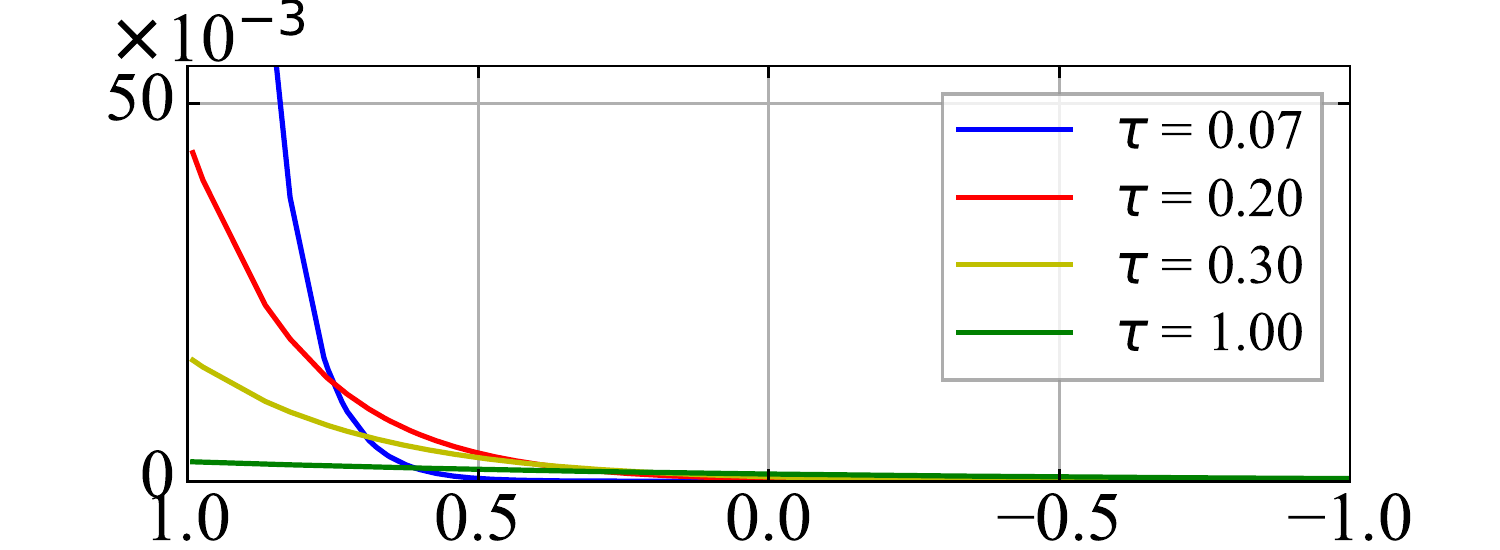}
    \caption{The gradient ratio $r_{i,j}$ with respect to different $s_{i,j}$. We sample the $s_{i,j}$ from a uniform distribution in $[-1,1]$. As we can see, with lower temperature, the contrastive loss tends to punish more on the hard negative samples. }
    \label{fig: r_distribution}
\end{figure}

\subsection{Gradients Analysis.}
We analyze the gradients with respect to positive samples and different negative samples. We will show that the magnitude of positive gradient is equal to the sum of negative gradients. The temperature controls the distribution of negative gradients. Smaller temperature tends to concentrate more on the nearest neighbours of the anchor point, which plays a role in controlling the hardness-aware sensitivity. Specifically, the gradients with respect to the positive similarity $s_{i,i}$ and the negative similarity $s_{i,j}$ ($j \neq i$) are formulated as:
\begin{equation}
\frac{\partial \mathcal{L}(x_i)}{\partial s_{i,i}} = 
- \frac{1}{\tau} \sum_{k \neq i} P_{i,k}, \quad 
\frac{\partial \mathcal{L}(x_i)}{\partial s_{i,j}} = 
\frac{1}{\tau} P_{i,j}
\label{form:gradient}
\end{equation}
From Eq \ref{form:gradient}, we have the following observations: (1) The gradients with respect to negative samples is proportional to the exponential term $exp(s_{i,j}/\tau)$, indicating that the contrastive loss is a hardness-aware loss function, which is different with the loss of Eq \ref{eq: lsimple} that gives all negative similarities the same magnitude of gradients. (2) The magnitude of gradient with respect to positive sample is equal to the sum of gradients with respect to all negative samples, i.e., $   (\sum_{k \neq i} | \frac{\partial L(x_i)}{\partial s_{i, k}}|) / |\frac{\partial L(x_i)}{\partial s_{i,i}}| = 1$, which can define a probabilistic distribution to help understand the role of temperature $\tau$.

\subsection{The Role of temperature}
The temperature plays a role in controlling the strength of penalties on hard negative samples. Specifically, we define $r_{i}(s_{i,j}) = | \frac{\partial L(x_i)}{\partial s_{i, j}} | / |\frac{\partial L(x_i)}{\partial s_{i,i}}|$, representing the relative penalty on negative sample $x_j$. We have:
\begin{equation}
r_{i}(s_{i,j}) = \frac{{\rm exp}(s_{i,j}/\tau)}{\sum_{k \neq i} {\rm exp}(s_{i,k}/\tau)}, \quad i \neq j
\end{equation}
which obeys the Boltzman distribution. As the temperature $\tau$ decreases, the entropy of the distribution $H(r_i)$ decreases strictly (the proof is in supplementary material). The distribution of $r_i$ becomes more sharp on the large similarity region, which gives large penalties to the samples closed to $x_i$. Fig \ref{fig: r_distribution} shows the relation of $r_i$ and $s_i$. From Fig \ref{fig: r_distribution}, we observe that the relative penalty concentrates more on the high similarity region as the temperature decreases, and the relative penalty distribution tends to be more uniform as the temperature increases, which tends to give all negative samples the same magnitude of penalties. Besides, the effective penalty interval become narrowed as the temperature decreases. Extremely small temperatures will cause the contrastive loss only concentrate on the nearest one or two samples, which will heavily degenerate the performance. In this paper, we keep the temperatures in a reasonable interval to avoid this situation.

Let us consider two extreme cases: $\tau \to 0^+$ and $\tau \to +\infty$. When $\tau \to 0^+$, we have the following approximation:
\begin{equation}
\begin{aligned}
&\lim\limits_{\tau \to 0^+}  - {\rm log}  \left[ \frac{ {\rm exp} (s_{i,i}/\tau)}{\sum_{k \neq i} {\rm exp} (s_{i,k}/\tau) + {\rm exp} (s_{i,i}/\tau)} \right] \\
 = & \lim\limits_{\tau \to 0^+}   + {\rm log} \left[ 1 + \sum_{k \neq i} {\rm exp}((s_{i,k}-s_{i,i})/\tau) \right] \\
 = & \lim\limits_{\tau \to 0^+}  + {\rm log} \left[ 1 + 
 \sum^k_{s_{i,k} \geqslant s_{i,i}} {\rm exp}((s_{i,k} - s_{i,i})/\tau) \right]\\
 = & \lim\limits_{\tau \to 0^+} \frac{1}{\tau}max[s_{max}-s_{i,i}, 0]
 \label{eq: app0}
\end{aligned}
\end{equation}
where $s_{max}$ is the maximum of the negative similarities. This shows that when $\tau \to 0^+$ the contrastive loss becomes a triplet loss with the margin of $0$, which only focuses on the nearest negative sample. When $\tau \to +\infty$, we approximate the contrastive learning as following:
\begin{equation}
\begin{aligned}
&\lim\limits_{\tau \to +\infty}  - {\rm log}  \left[ \frac{ {\rm exp} (s_{i,i}/\tau)}{\sum_{k \neq i} {\rm exp} (s_{i,k}/\tau) + {\rm exp} (s_{i,i}/\tau)} \right] \\
= & \lim\limits_{\tau \to +\infty} - \frac{1}{\tau} s_{i,i} + {\rm log}
\sum_{k} {\rm exp}(s_{i,k}/\tau)
 \\
= & \lim\limits_{\tau \to +\infty} -\frac{1}{\tau} s_{i,i} + \frac{1}{N} \sum_{k} {\rm exp}(s_{i,k}/\tau) - 1 + {\rm log} N \\
= & \lim\limits_{\tau \to +\infty} -\frac{N-1}{N\tau} s_{i,i} + \frac{1}{N\tau} \sum_{k \neq i} s_{i,k} + {\rm log}N
\label{eq: appinfty}
\end{aligned}
\end{equation}
We use the Taylor expansion of ${\rm log}(1+x)$ and ${\rm exp}(x)$ and omit the second or higher order infinitesimal terms. The above approximation of contrastive loss is equivalent to the simple contrastive loss $\mathcal{L}_{{\rm simple}}$, which shows that the simple contrastive loss is a special case of the softmax-based contrastive loss by approaching the temperature to $+\infty$.

We also conduct experiments to study the behaviours of the two extreme cases. Specifically, using the objective of Eq \ref{eq: app0}, the model can not learn any useful information. Using Eq \ref{eq: appinfty} as the objective, the performances on downstream tasks are inferior to the models trained with the ordinary contrastive loss by a relative large margin. However, combining the loss of Eq \ref{eq: appinfty} with an explicit hard negative sampling strategy, the model will achieve competitive downstream results, which shows the importance of the hardness-aware property of the contrastive loss.

\subsection{Explicit Hard Negative Sampling}
In this subsection, we study a more straightforward hard negative sampling strategy which truncates the gradients with respect to the uninformative negative samples. Specifically, given an upper $\alpha$ quantile $s_\alpha^{(i)}$ for the anchor sample $x_i$, we define the \textit{informative interval} as $[s_\alpha^{(i)}, 1.0]$, and the \textit{uninformative interval} as $[-1.0, s_\alpha^{(i)}]$. We force the gradient ratio of $s_{i,j}$ which resides in the uninformative interval to 0, i.e., $r_{i}(s_{i,j}) = 0$ for $s_{i,j} < s_\alpha^{(i)}$, and the gradient ratio of $x_l$ residing in the informative interval as:

\begin{equation}
r_{i}(s_{i,l}) = \frac{{\rm exp}(s_{i,l}/\tau)}{\sum_{s_{i,k} \geqslant s_{\alpha}^{(i)} } {\rm exp}(s_{i,k}/\tau)}, \quad l \neq i
\end{equation}
The above operation squeezes the negative gradients from the uninformative interval to the informative interval. The corresponding hard contrastive loss is:
\begin{equation}
\mathcal{L}_{{\rm hard}}(x_i) = - {\rm log}  \frac{ {\rm exp} (s_{i,i}/\tau)} {\sum_{s_{i,k} \geqslant s_{\alpha}^{(i)} } {\rm exp} (s_{i,k}/\tau) + {\rm exp} (s_{i,i}/\tau)} 
\label{eq: hardloss}
\end{equation}

The $\mathcal{L}_{{\rm hard}}$ only penalizes the informative hard negative samples. The hard contrastive loss acts on hard negative samples in two ways: an explicit way that chooses the top $K$ nearest negative samples and an implicit way by the hardness-aware property. Using the same temperature with the contrastive loss of Eq \ref{equation: conloss}, the hard contrastive loss usually generate more uniform embedding distribution, and it is beneficial to choose relative large temperatures. Besides, with this explicit hard negative sampling strategy, we show that the current popular contrastive loss of Eq \ref{equation: conloss} can be replaced by the simple form of Eq \ref{eq: lsimple}, with similar or even better performances on downstream tasks. Note that we are not the first to propose the idea of the above hard contrastive loss. LocalAggregation proposed by Zhuang et al\cite{zhuang2019local} have used the above hard negative mining strategy. In this paper, we will concentrate on analyzing the behaviour of this contrastive loss.


\section{Uniformity-Tolerance Dilemma} \label{section:utd}
In this section, we study two properties: uniformity of the embedding distribution and the tolerance to semantically similar samples. The two properties are both important to the feature quality. 

\begin{figure}[t]
\centering
\includegraphics[width=0.95\linewidth]{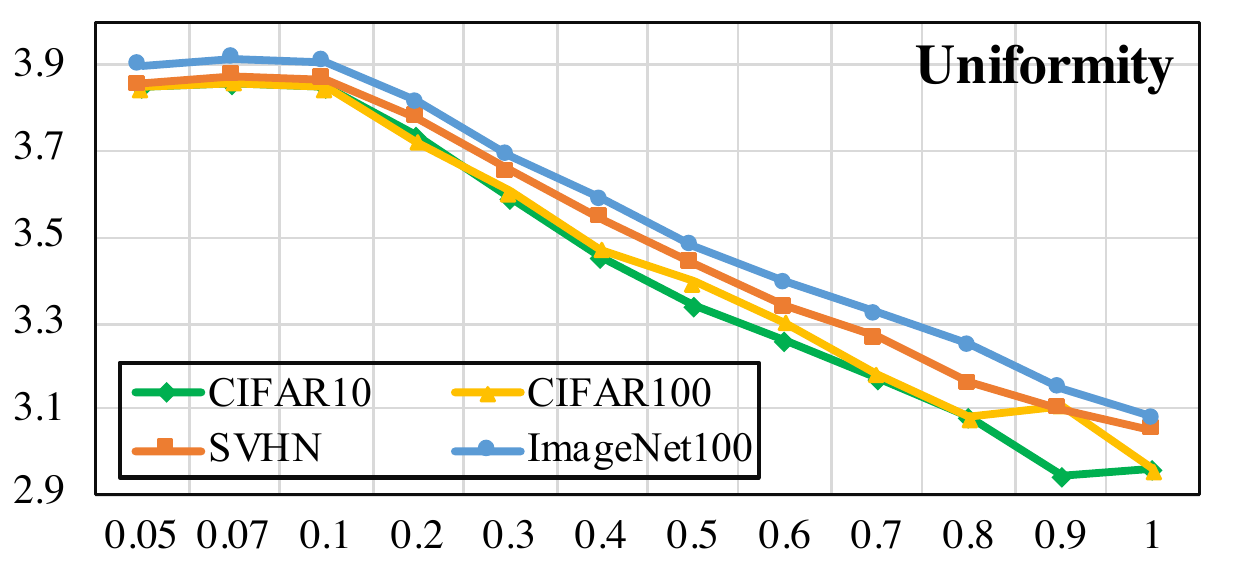}
\caption{Uniformity of embedding distribution trained with different temperature on CIFAR10, CIFAR100 and SVHN. The x axis represents different temperature, and y axis represents $-\mathcal{L}_{{\rm uniformity}}$. Large value means the distribution is more uniform.}
\label{fig: uniformity}
\end{figure}
\begin{figure}[t]
\centering
\includegraphics[width=0.95\linewidth]{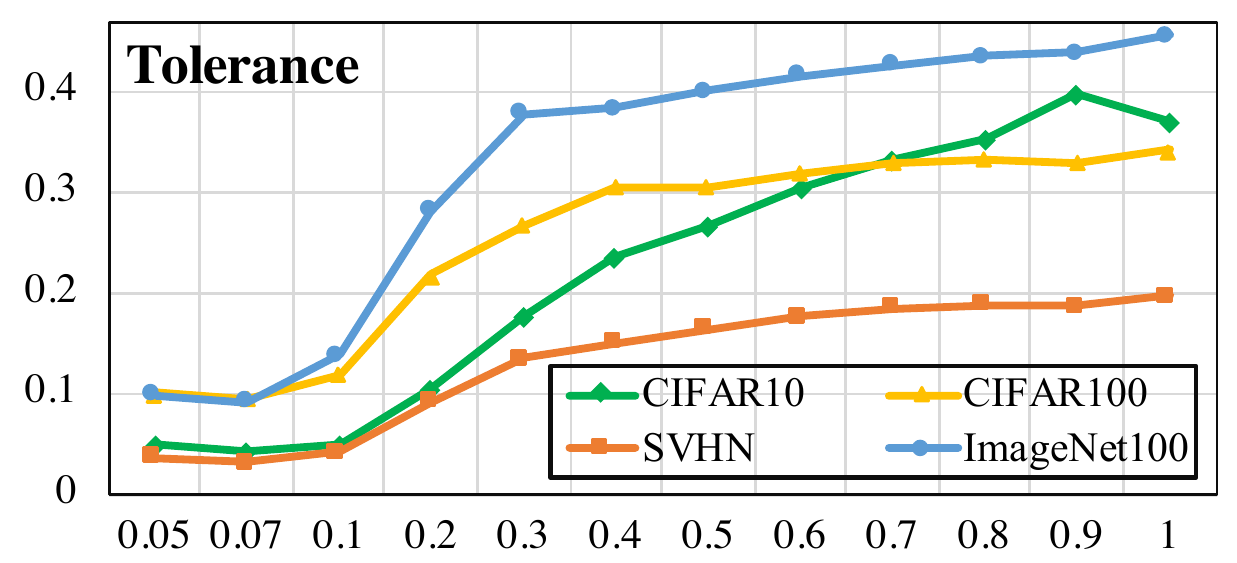}
\caption{Measurement of tolerance on models trained on CIFAR10, CIFAR100 and SVHN. The x axis represents different temperatures, and y axis represents the tolerance to samples with the same category. Large value means the model is more tolerant to semantically consistent samples.}
\label{fig: tolerance}
\end{figure}

\subsection{Embedding Uniformity}
In \cite{wang2020hypersphere}, the authors find that the uniformity is a significant property in contrastive learning. The contrastive loss can be distangled to two parts, which encourages the positive features to be aligned and the embeddings to match a uniform distribution in a hypersphere. In this part, we will explore the relation between the local separation and the uniformity of embeddings. To this end, we incorporate the uniformity metric proposed by \cite{wang2020hypersphere}, which is based on a gaussian potential kernel:
\begin{equation}
\mathcal{L}_{{\rm uniformity}}(f;t) = {\rm log} \mathop{\mathbb{E}}\limits_{x,y \sim p_{data}}\left[ e^{-t||f(x)-f(y)||_2^2} \right]
\label{eq: uniformity}
\end{equation}

We calculate $\mathcal{L}_{{\rm uniformity}}$ on models trained with different temperatures to control different levels of local separation. We trained different models on CIFAR10, CIFAR100, SVHN and ImageNet100. Fig \ref{fig: uniformity} shows the tendency. As the temperature increases, the embedding distribution tends to be less uniform (In Fig \ref{fig: uniformity}, the y-axis represents the $-\mathcal{L}_{{\rm uniformity}}$). And when $\tau$ is small, the embedding distribution is closer to a uniform distribution. This can be explained as follows: when the temperature is small, the contrastive loss tends to separate the positive samples close to the anchor sample, which makes the local distribution be sparse. With all samples are trained, the embedding space tends to make the neighbour of each point be sparse, and the distribution tends to be more uniform. For the hard contrastive loss, the situation is illustrated in Fig \ref{fig: uniformity_hard}. With the hard contrastive loss as objective, the distribution tends to be more uniform. Besides, the uniformity keeps relative stable with the change of temperature compared with the ordinary contrastive loss. The explicit hard negative sampling weakened the effect of the temperature to control the hardness-aware property.

\subsection{Tolerance to Potential Positive Samples}
The objective of contrastive learning is to learn the augmentation alignment and instance discriminative embedding. The contrastive loss has no constraint to the distribution of the negative samples. However, with the help of heavy augmentation and strong abstraction ability of deep convolutional neural networks, the negative distribution reflects some extent of semantics, which is illustrated in Fig \ref{fig: concept} (a). However, from the above section we have recognized that when the temperature $\tau$ is very small, the penalties to the nearest neighbours will be strengthened, which will push the semantically similar samples strongly to break the semantic structure of the embedding distribution. To explain the phenomenon in a quantitative manner, we measure the tolerance to the semantically consistent samples using the mean similarities of samples belong to the same class, which is formulated as:
\begin{equation}
T = \mathop{\mathbb{E}}\limits_{x,y \sim p_{data}} \left[(f(x)^Tf(y)) \cdot I_{l(x)=l(y)}\right]
\label{eq: tolerance}
\end{equation}
where $l(x)$ represents the supervised label of image $x$. $I_{l(x)=l(y)}$ is an indicator function, having the value of 1 for $l(x)=l(y)$ and the value of 0 for $l(x) \neq l(y)$. Fig \ref{fig: tolerance} shows the tolerance with respect to different temperatures on CIFAR10 and CIFAR100. We could see that the tolerance is positively related to the temperature $\tau$. However, the tolerance can not directly reflect the feature quality. For example, when all the samples reside in a single point of the hypersphere, then the tolerance is maximized, while the feature quality is bad. The tolerance reflects the local density of semantically related samples. An ideal model should be both locally clustered and globally uniform. 

The contrastive loss meets a uniformity-tolerance dilemma. On the one hand, we hope to decrease the temperature $\tau$ to increase the uniformity of the embedding distribution, on the other hand, we hope to increase the temperature to make the embedding space tolerant to the similar samples. For the ordinary contrastive loss, it is a compromise to choose the appropriate temperature to balance both the embedding uniformity and the tolerance to semantically similar samples. The dilemma is caused by the inherent defect of unsupervised contrastive loss that it pushes all different instances ignoring their semantical relations. 

\begin{figure}[t]
\centering
\includegraphics[width=1.0\linewidth]{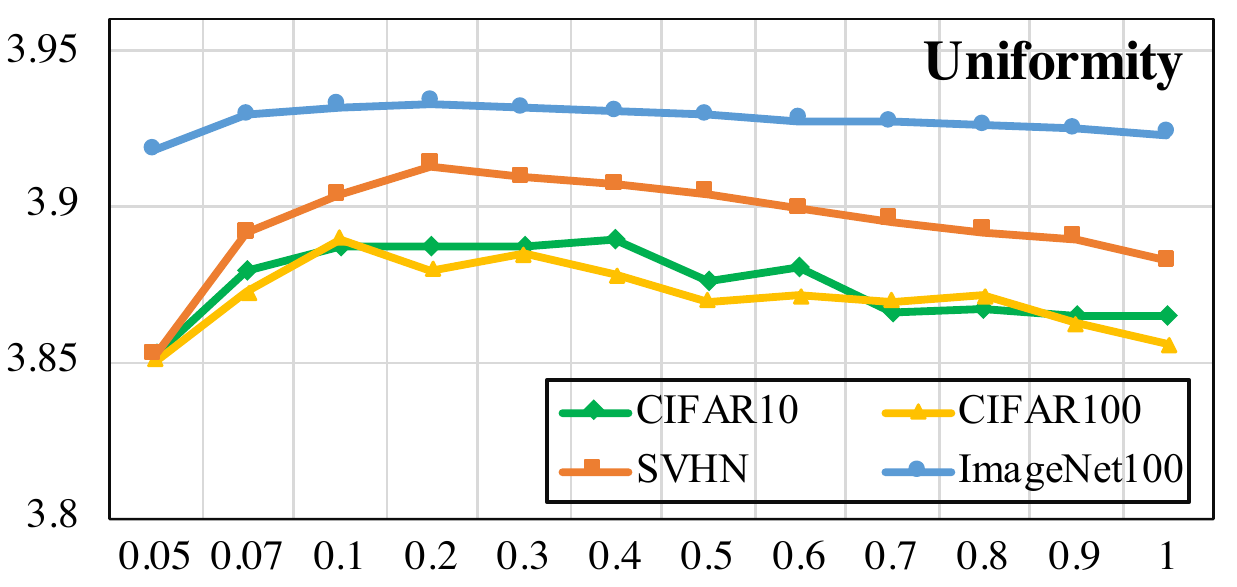}
\caption{Uniformity of embedding distribution trained with hard contrastive loss $\mathcal{L}_{{\rm hard}}$ on the three datasets. The x axis represents different temperature, and y axis represents $-\mathcal{L}_{{\rm uniformity}}$. Large value means the distribution is more uniform.}
\label{fig: uniformity_hard}
\end{figure}
\begin{figure}[t]
\centering
\includegraphics[width=1.0\linewidth]{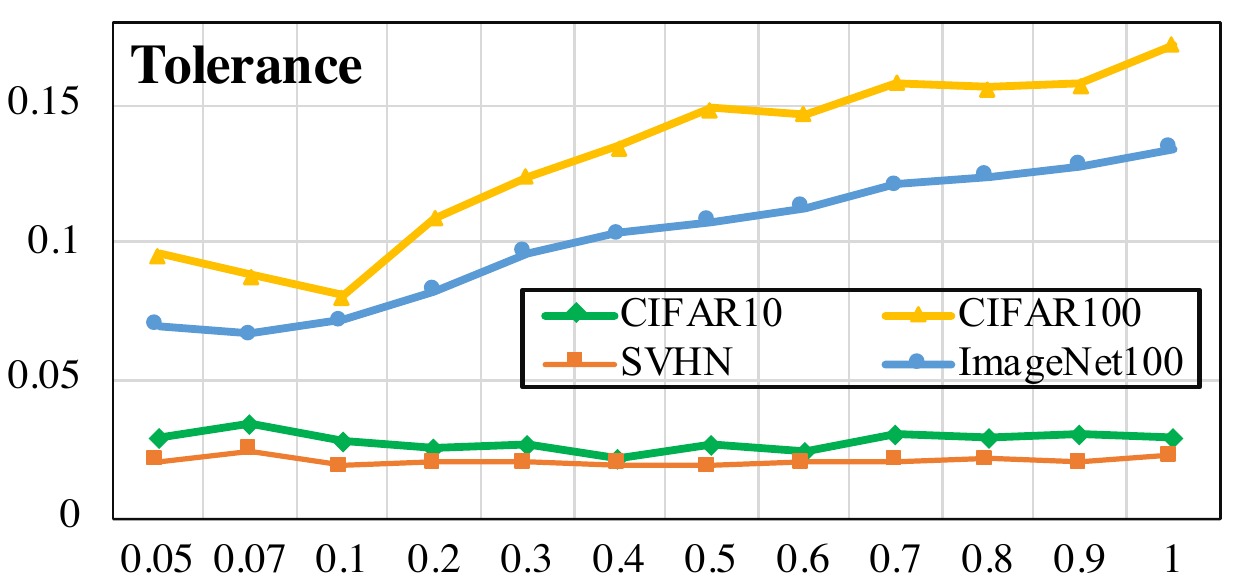}
\caption{Measurement of tolerance on models trained on the three datasets with hard contrastive loss $\mathcal{L}_{{\rm hard}}$. The x axis represents different temperatures, and y axis represents the tolerance to samples with the same category. Large value means the model is more tolerant to semantically consistent samples.}
\label{fig: tolerance_hard}
\end{figure}

\begin{figure*}[t]
\centering
\includegraphics[width=1.0\linewidth]{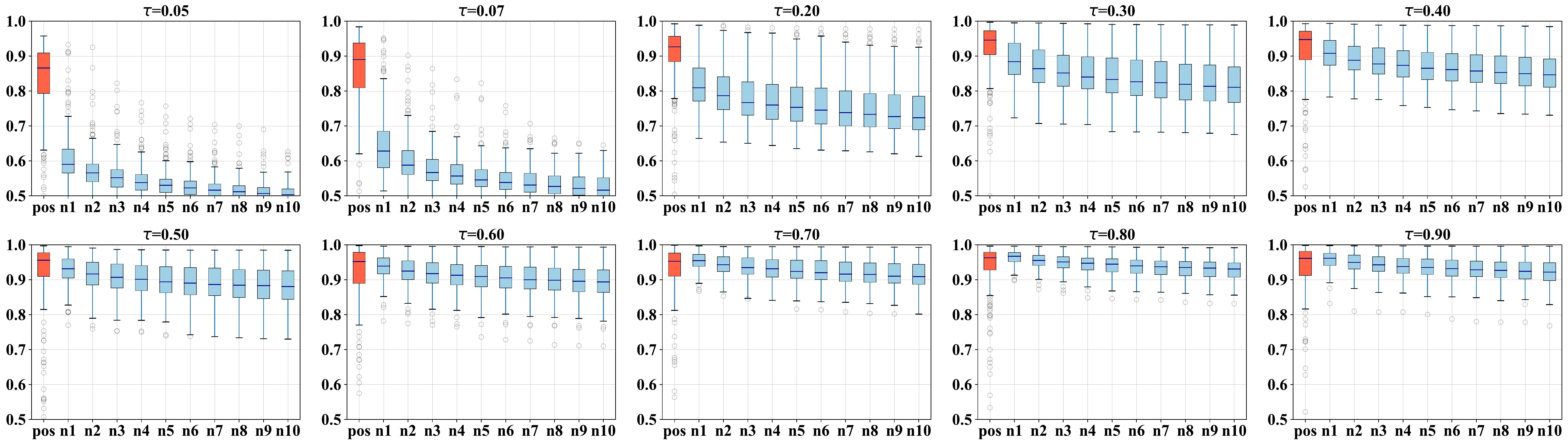}
\caption{We display the similarity distribution of positive samples and the top-10 nearest negative samples that are marked as 'pos' and 'ni' for the i-th nearest neighbour. All models are trained on CIFAR100. For models trained on other datasets, they present the same pattern with the above figure, and we display them in the supplementary material. }
\label{fig: exp1}
\end{figure*}

Fig \ref{fig: uniformity_hard} and Fig \ref{fig: tolerance_hard} show the measurement of the embedding uniformity and the tolerance to samples in the same categories. We will see that the embedding distribution produced by hard contrastive loss is more uniform than the ordinary contrastive loss. This is caused by the increased gradients on the informative samples. Correspondingly, the tolerance to potential positive samples is decreased compared with the ordinary contrastive loss. However, the decrease of tolerance is caused by the increased uniformity, i.e., similarities with the samples in different categories are also decreased. 

The hard contrastive loss deals better with the uniformity-tolerance dilemma. As we can see from Fig \ref{fig: uniformity_hard} and Fig \ref{fig: tolerance_hard}, the uniformity keeps relative stable compared with the ordinary contrastive loss (from Fig \ref{fig: uniformity}). Relative large temperature can help be more tolerant to the potential positive samples without decreasing too much uniformity. We consider this is because the explicit hard negative sampling strategy is very effective for generating uniform embedding distribution. 

\begin{figure*}[t]
\centering
\includegraphics[width=1.0\linewidth]{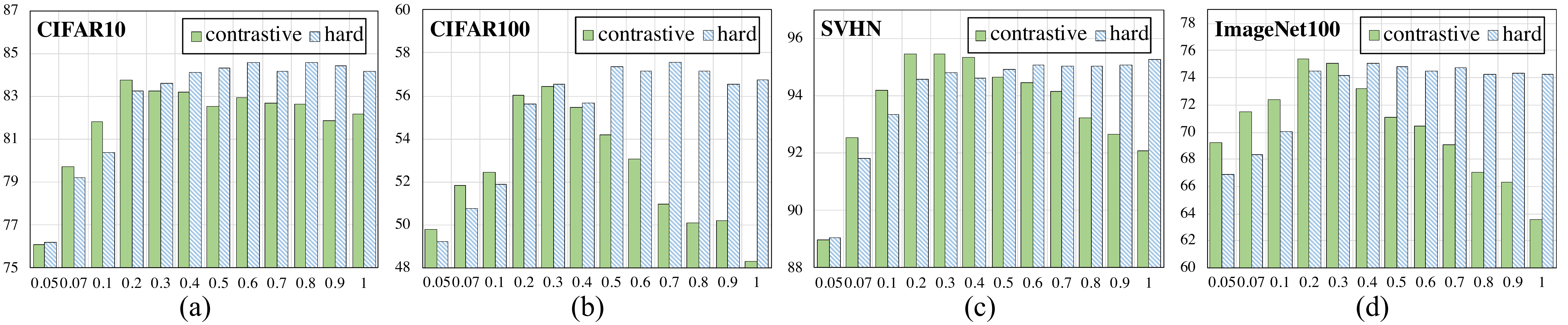}
\caption{Performance comparison of models trained with different temperatures. For CIFAR10, CIFAR100 and SVHN, the backbone network is ResNet-18, and for ImageNet, the backbone network is ResNet-50. After the pretraining stage, we freeze all convolutional layers and add a linear layer. We report 1-crop top-1 accuracy for all models.}
\label{fig: results}
\end{figure*}

\section{Results}

\subsection{Experiment Details}

\textbf{Pretraining.} We conduct experiments on CIFAR10, CIFAR100 \cite{Krizhevsky2009LearningML}, SVHN \cite{netzer2011reading} and ImageNet100 \cite{deng2009imagenet}. The labels of the ImageNet100 are listed in the supplementary material. For the pretraining stage, we use resnet18 \cite{he2016deep} with a minor modification (change the size of the first convolutional kernel as $3 \times 3$ to adapt to $32 \times 32$ input) as the backbone on CIFAR10, CIFAR100 and SVHN, and we use resnet50 \cite{he2016deep} as the backbone on ImageNet100. For CIFAR10, CIFAR100 and SVHN, the augmentations follow \cite{wu2018unsupervised}: a $32 \times 32$ pixel crop is taken from a randomly resized image, and then undergoes random color jittering, random horizontal flip, and random gray scale conversion. For ImageNet-100, we follows \cite{chen2020simple} to add a random gaussian blur operation. To save the negative features, we follow \cite{wu2018unsupervised} to create a memory bank which records the exponential moving average of the learned features. We use SGD as our optimizer. The SGD weight decay is 5e-4 for CIFAR10, CIFAR100 and SVHN, and 1e-4 for ImageNet100. The SGD momentum is set to 0.9. For the hard contrastive loss, the $\alpha$ is set to 0.0819, 0.0819, 0.0315 and 0.034 for CIFAR10, CIFAR100, SVHN, and ImageNet100 (4095 negative samples). We train all models for 200 epochs with the learning rate multiplied by 0.1 at 160 and 190 epochs. We set an initial learning rate as 0.03, with a mini-batch size of 128. 

\textbf{Evaluation.} We validate the performance of the pretrained models on linear classification models. Specifically, we train the linear layer for 100 epochs, with all convolutional layers frozen. We set an initial learning rate of 30.0, which is multiplied by 0.2 at 40, 60 and 80 epochs, and use SGD optimizer with weight decay of 0.

\subsection{Local Separation}
In this subsection, we evaluate the effect of the temperature. First, we try to figure out if the temperature accurately controls the strength of penalties on hard negative samples, furthermore, the extent of local separation. Specifically, we calculate $s_{i,j}$ for all point $x_j$ given an anchor sample $x_i$, and then take an average over all anchor samples. We sort the similarities in a descending order and observe the distribution of the positive similarities $s_{i,i}$ and ten largest negative similarities that for all $s_{i,l} \in Top_{10}(\{s_{i,j}| \forall j \neq i \})$. We calculate these positive and negative similarities with the models trained on CIFAR100 and display them in Fig \ref{fig: exp1} (It is the same pattern when we calculate them on other datasets displayed in supplementary material). From Fig \ref{fig: exp1}, we observe that: (1) As the $\tau$ decreases, the gap between positive samples and other confusing negative samples are larger, i.e., the positive and negative samples are more separable. (2) As $\tau$ increases, the positive similarities tend to be closer to 1. Observation (1) shows that small temperature indeed tends to push the hard negative samples more significantly, as indicated in the Fig \ref{fig: r_distribution}, that small temperature makes the distribution of $r_i$ more sharply and concentrate most the penalties on the hardest negative samples (nearest neighbours). As the temperature increases, the positive samples and some confusing negative samples are likely to be less discriminative, and the relative penalties distribution $r_i$ tends to be more uniform to concentrate less on the hard negative samples. Observation (2) shows that as the temperature increases, the positive samples are more aligned, and the model tends to learn features more invariant to the data augmentations. We explain that the observation (2) is also caused by the role of temperature. For example, when the temperature is small, the contrastive loss punishes the hardest samples which are likely to share the similar content as the augmentations. Punishing these similar negative samples away significantly will make the objective of making positive samples alignment puzzling.

\begin{table*}
    \centering
    \begin{tabular}{l|c|c|c|c|c|c|c|c|c|c|c}
      \noalign{\hrule height 1pt}
      \multirow{2}{*}{\textbf{Dataset}} &
      \multirow{2}{*}{\textbf{Result}} &
      \multicolumn{4}{c|}{\textbf{Contrastive}} &
      \multirow{2}{*}{\textbf{Simple}} &
      \multicolumn{4}{c|}{\textbf{HardContrastive}} &
      \multirow{2}{*}{\textbf{HardSimple}} \\
      \cline{3-6} \cline{8-11}
      & & 0.07 & 0.3 & 0.7 & 1.0 & & 0.07 & 0.3 & 0.7 & 1.0 & \\
      \noalign{\hrule height 1pt}
      \multirow{3}{*} {\textbf{CIFAR10}} & accuracy & 79.75 & 83.27 & 82.69 & 82.21 & 74.83 & 79.2 & 83.63 & 84.19 & 84.19 &   84.84               \\
      & uniformity                       & 3.86  & 3.60 & 3.17 & 2.96 & 1.68  & 3.88 & 3.89 & 3.87 & 3.86 & 3.85\\
      & tolerance                        & 0.04  & 0.178 & 0.333 & 0.372 & 0.61  & 0.034 & 0.0267 & 0.030 & 0.030 & 0.030\\
        \hline
      \multirow{3}{*} {\textbf{CIFAR100}} & accuracy & 51.82 & 56.44 & 50.99 & 48.33 & 39.31 &  50.77 & 56.55 & 57.54 & 56.77 &   55.71               \\
      & uniformity                       & 3.86  & 3.60 & 3.18 & 2.96 & 2.12  & 3.87 & 3.88 & 3.87 & 3.86 & 3.86\\
      & tolerance                        & 0.10  & 0.269 & 0.331 & 0.343 & 0.39  & 0.088 & 0.124 & 0.158 & 0.172 & 0.174\\
        \hline
        \multirow{3}{*} {\textbf{SVHN}} & accuracy  & 92.55 & 95.47 & 94.17 & 92.07 & 70.83 & 91.82 & 94.79 & 95.02  & 95.26 &   94.99     \\
      & uniformity                       & 3.88  & 3.65 & 3.27 & 3.05 & 1.50  & 3.89 & 3.91 & 3.90 & 3.88 & 3.85\\
      & tolerance                        & 0.032  & 0.137 & 0.186 & 0.197 & 0.074 & 0.025 & 0.021 & 0.021 & 0.023 & 0.026\\
        \hline
        \multirow{3}{*} {\textbf{ImageNet100}} & accuracy  & 71.53 & 75.10 & 69.03 & 63.57 & 48.09 & 68.33 & 74.21 & 74.70 & 74.28  & 74.31     \\
      & uniformity                       & 3.917 & 3.693 & 3.323 & 3.08 & 1.742 & 3.929 & 3.932 & 3.927 & 3.923 & 3.917\\
      & tolerance                        & 0.093 & 0.380 & 0.427 & 0.456 & 0.528 & 0.067 & 0.096 & 0.121 & 0.134 & 0.157\\
      \noalign{\hrule height 1pt}
    \end{tabular}
    \vspace*{2mm}
    \caption{We report the accuracy of linear classification on CIFAR10, CIFAR100 and SVHN, including models trained with the ordinary contrastive loss, simple contrastive loss, hard contrastive loss and hard simple contrastive loss. For models trained on ordinary contrastive loss and hard contrastive loss, we select several representative temperatures. More results are shown in the supplementary material.}
    \label{table: results}
\end{table*}

\subsection{Feature Quality}
We evaluate the performance of the contrastive models with different settings on cifar10, cifar100, SVHN and ImageNet100. Fig \ref{fig: results} shows the performances of linear classification on the four datasets respectively. For the models trained with ordinary contrastive loss (Eq \ref{equation: conloss}), the performance tends to present a reverse-U shape. The models achieve the best performance when the temperature is 0.2 or 0.3. Models with small or large temperature achieve suboptimal performances. The results indicate that it is a compromise between uniformity and the tolerance. Models with small temperature tend to generate uniform embedding distribution, while they break the underlying semantic structure because they give large magnitudes of penalties to the closeness of potential positive samples. It is harmful to concentrate on the hardest negative samples due to they are very likely to be the samples whose semantic properties are very similar to the anchor point. On the other hand, models with large temperature tends to be more tolerant to the semantically consistent samples, while they may generate embeddings with not enough uniformity. Table \ref{table: results} shows the numerical results, from which we can see that although the tolerance increases as the temperature increases, the uniformity decreases. This indicates that the embeddings tend to reside in a crowd region on the hypersphere. For the models trained with the hard contrastive loss (Eq \ref{eq: hardloss}), the above uniformity-tolerance dilemma is alleviated. From Fig \ref{fig: results}, we observe that the models trained with hard contrastive loss achieve better results when the temperatures are large enough. This is because the uniformity is guaranteed by the explicit hard negative mining, which is reflected in Fig \ref{fig: uniformity_hard}. 

\subsection{Uniformity and Tolerance}
To measure the uniformity of embedding distribution and the tolerance to the semantically similar samples, we use Eq \ref{eq: uniformity} and \ref{eq: tolerance} as the measurement of those two properties. The experiments are conducted on CIFAR10, CIFAR100, SVHN and ImageNet100 respectively. Fig \ref{fig: uniformity} and Fig \ref{fig: tolerance} show the uniformity and tolerance of models trained with ordinary contrastive loss. Fig \ref{fig: uniformity_hard} and Fig \ref{fig: tolerance_hard} show the uniformity and tolerance of models trained with the hard contrastive loss. Detailed analysis is presented in Section \ref{section:utd}. Concrete numerical values are present in Table \ref{table: results} for some representative models, all results are listed in supplementary material.

\subsection{Substitution of Contrastive Loss}
We have claimed that the hardness-aware property is a key property to the success of contrastive loss. In this part, we will show that with explicit hard negative sampling strategy, the softmax-based contrastive loss of Eq \ref{equation: conloss} is not necessary, and a simple contrastive loss of Eq \ref{eq: lsimple} works pretty well and achieve competitive results. Table \ref{table: results} shows the results. Concretely, we use the simple contrastive loss of Eq \ref{eq: lsimple} as objective, which is equivalent to the extreme case as $\tau$ approaches $+\infty$, and is marked as Simple in Table \ref{table: results}. Besides, we also trained models with a hard simple contrastive loss, using the nearest 4095 features as negative samples, which is marked as HardSimple in Table \ref{table: results}. Without the hardness-aware property, the learned models with $\mathcal{L}_{{\rm simple}}$ perform much worse than models trained with ordinary contrastive loss (74.83 vs 83.27 on CIFAR10, 39.31 vs 56.44 on CIFAR100, 70.83 vs 95.47 on SVHN, 48.09 vs 75.10 on ImageNet100). However, when the negative samples of the $\mathcal{L}_{{\rm simple}}$ are drawn from the nearest neighbours, the trained models achieve competitive results on all three datasets. This shows that the hardness-aware property is the core to the success of the contrastive loss.

\section{Conclusion}
In this paper, we try to understand the behaviour of the unsupervised contrastive loss. We show that the contrastive loss is a hardness-aware loss function, and the hardness-aware property is significant to the success of the contrastive loss. Besides, the temperature plays a key role in controlling the local separation and global uniformity of the embedding distributions. With the temperature as a proxy, we have studied the uniformity-tolerance dilemma, which is a challenge met by the unsupervised contrastive learning. We believe the uniformity-tolerance dilemma can be addressed by explicitly modeling the relation between different instances. We hope our work can inspire researchers to explore such algorithm to address the uniformity-tolerance dilemma.

\section{Acknowledgments}
This work was supported in part by the National Natural Science Foundation Project under Grant 62025304 and in part by the Seed Fund of Tsinghua University (Department of Computer Science and Technology)-Siemens Ltd., China Joint Research Center for Industrial Intelligence and Internet of Things.

{\small
\bibliographystyle{ieee_fullname}
\bibliography{CVPR2021_8484_FINAL}
}

\end{document}


\title{Supplementary Material: Understanding the Behaviour of Contrastive Loss}

\maketitle


\section{Introduction}
In this supplementary material, we list some detailed results of our paper including: \textbf{(1)} The proof of monotonicity of entropy with respect to temperature coeffecient $\tau$. \textbf{(2)} All numerical results of different models trained with contrastive loss are shown in Table 1, and the results of different models trained with hard contrastive loss are shown in Table 2. We train these models using different temperatures ranging from $0.05$ to $1.0$ on CIFAR10, CIFAR100, SVHN and ImageNet100. \textbf{(3)} For the ImageNet100 dataset, we list the 100 labels of ImageNet100 which is shown in Table 3. \textbf{(4)}. The illustrations of the local separation on different datasets. In our paper, we have shown the local separation property on CIFAR100 dataset, and have found that the local separation property on all datasets is similar. Figure 1-3 show the local separation on CIFAR10, CIFAR100 and ImageNet100. Fig 4-7 show the local separation property of the hard contrastive loss on the four datasets.

\section{Proof in Sec3.2}
In Sec3.2 (The role of temperature), we have stated that the entropy $H(r_i)$ increases strictly as the temperature increases. In this part, we prove this statement. Specifically, given a distribution $r_i(s_{i,j})$ as:
\begin{equation}
    r_{i}(s_{i,j}) = \frac{{\rm exp}(s_{i,j}/\tau)}{\sum_{k \neq i} {\rm exp}(s_{i,k}/\tau)}, \quad i \neq j
\end{equation}
, what we hope to prove is that when other variables including $s_{i,j}$ and $s_{i,k}$ keep invariant, the entropy is $H(r_i)$ is monotonically increasing (Except for the special case when all $s_{i,k}$ is equal, which makes the $r_i$ be a uniform distribution).
For simplicity, let:
\begin{equation}
P_l = {\rm exp}(s_{i,l}/\tau)
\end{equation}
We then re-write the entropy $H$ using the above symbol as follows:
\begin{equation}
\begin{aligned}
    H(r_i) &= - \sum_{j \neq i} r_i(s_{i,j}) \cdot {\rm log}(r_i(s_{i,j})) \\
           &= - \frac{\sum_{j \neq i} P_j\cdot {\rm log}(P_j)}{\sum_{j \neq i} P_j} + {\rm log}(\sum_{j \neq i} P_j)
\end{aligned}
\end{equation}
Next, we calculate the gradients of $H$ with respect to $P_l$ for any $l \neq i$ as follows:
\begin{equation}
\begin{aligned}
 \frac{\partial{H}}{\partial{P_l}} & = - \frac{{\rm log} P_l}{\sum_{j \neq i} P_j} + \frac{\sum_{j \neq i} P_j {\rm log} P_j}{(\sum_{j \neq i} P_j)^2} 
\end{aligned}
\end{equation}
and the gradient of $P_l$ with respect to $1/\tau$:
\begin{equation}
\begin{aligned}
 \frac{\partial{P_l}}{\partial{1/\tau}} = \tau P_l \cdot log(P_l)
\end{aligned}
\end{equation}

We have computed the gradient of $H$ with respect to $P_l$ and the gradient of $P_l$ with respect to $1/\tau$. Using the chain rule, we can calculate the gradient of $H$ with respect to $1/\tau$ is as follows:

\begin{equation}
\begin{aligned}
    &\frac{\partial{H}}{\partial{1/\tau}} = \sum_l \frac{\partial{H}}{\partial{P_l}} \cdot \frac{\partial{P_l}}{\partial{1/\tau}} \\
    &= \tau \cdot \frac{(\sum_{l \neq i} P_l \cdot {\rm log}(P_l))^2 - \sum_{l \neq i} P_l \sum_{l \neq i} P_l {\rm log}^2(P_l)}{(\sum_{l \neq i} P_l)^2} \\
\end{aligned}
\end{equation}

Up to now, we have calculated the gradient of $H$ with respect to $1/\tau$ as the above equation, which only consists of $\tau$ and the proposed symbol $P_l$.
Notice that $P_l > 0$, we can apply the Cauchy inequality to the numerator part of the above equation. We have:

\begin{equation}
\begin{aligned}
    \sum_{l \neq i} P_l \sum_{l \neq i} P_l {\rm log}^2(P_l) & = \sum_{l \neq i} \sqrt{P_l}^2 \sum_{l \neq i} (\sqrt{P_l} \cdot {\rm log}(P_l))^2 \\
   & \geqslant (\sum_{l \neq i} P_l \cdot {\rm log}(P_l))^2
\end{aligned}
\end{equation}
,such that $\partial H/\partial(1/\tau) \leqslant 0$. In another word, the entropy is monotonically increasing as the $\tau$ increases. Furthermore, we notice that the equality of the Cauchy inequality is satisfied only if all $P_j$ is equal, which is almost impossible to satisfy in the learning process. 

\section{Results}
We list detailed experiment results and the chosen ImageNet100 labels as the following tables and figures.

\begin{table*}[t]
\newcommand{\tabincell}[2]{\begin{tabular}{@{}#1@{}}#2\end{tabular}}
	\centering
	\begin{tabular}{|l|c|c|c|c|c|c|c|c|c|c|c|c|c|}
        \noalign{\hrule height 1pt}
		\textbf{dataset} & 0.05 & 0.07 & 0.10 & 0.20 & 0.30 & 0.40 & 0.50 & 0.60 & 0.70 & 0.80 & 0.90 & 1.0  \\
		\hline
        CIFAR10 & 76.10& 79.75& 81.82& 83.78& 83.27&83.22& 82.54 & 82.97&82.69 & 82.67& 81.97&82.21 \\
        \hline
        CIFAR100  & 49.80& 51.82& 52.46&56.05 & 56.44&55.47 &54.17 &53.05 & 50.99& 50.08& 50.21& 48.33    \\
        \hline
        SVHN   & 88.96 & 92.55 & 94.21 & 95.46 & 95.47 & 95.36 & 94.66 & 94.47 & 94.17 & 93,22 & 92.66 & 92.07 \\
        \hline
        ImageNet100 & 63.91 & 71.53 & 74.59 & 75.41 & 75.10 & 72.98 & 71.10 & 70.47 & 69.03 & 67.91 & 65.93 & 65.49 \\
        \noalign{\hrule height 1pt}
	\end{tabular}
    \vspace{0.2cm}
	\caption{All results of different models trained with the ordinary contrastive loss. We test all models on a linear classification task, which freezes all convolutional layers and adds a linear layer on top of the last convolutional layer. We evaluate the above contrastive models on CIFAR10, CIFAR100, SVHN and ImageNet100 respectively.}
    \vspace{-1.2cm}
    \label{numerical_results}
\end{table*}

\begin{table*}[t]
	\centering
	\label{transfer}
	\begin{tabular}{|l|c|c|c|c|c|c|c|c|c|c|c|c|c|}
        \noalign{\hrule height 1pt}
		\textbf{dataset} & 0.05 & 0.07 & 0.10 & 0.20 & 0.30 & 0.40 & 0.50 & 0.60 & 0.70 & 0.80 & 0.90 & 1.0  \\
        \hline
        CIFAR10 & 76.22& 79.20&80.44 &83.28 & 83.63&84.14 &84.31          & 84.60 & 84.19&84.60&84.45 &84.19 \\
        \hline
        CIFAR100 & 49.21&50.77 & 51.91 &55.61& 56.55&55.66 &57.37           & 57.17&57.54 &57.15 &56.53 &56.77 \\

        \hline
        SVHN & 89.07 & 91.82 & 93.37 & 94.58 & 94.79 & 94.62 & 94.93 & 95.06 & 95.02 & 95.03 & 95.08 & 95.26 \\
        \hline
        ImageNet100 & 61.54 & 68.33 & 72.10 & 74.53 & 74.21 & 75.04 & 74.84 & 74.46 & 74.70 & 74.28 & 74.74 & 73.78\\
        \noalign{\hrule height 1pt}
	\end{tabular}
    \vspace{0.2cm}
	\caption{All results of different models trained with the hard contrastive loss. We test all models on a linear classification task, which freezes all convolutional layers and adds a linear layer on top of the last convolutional layer. We evaluate the above hard contrastive models on CIFAR10, CIFAR100, SVHN and ImageNet100 respectively.}
\end{table*}
\vspace{-1.2cm}


\begin{table*}[t]
	\centering
	\label{im100label}
	\footnotesize
	\begin{tabular}{|c|c|c|c|c|c|c|c|c|c|}
        \noalign{\hrule height 1pt}
        n01558993& n01692333& n01729322& n01735189& n01749939& n01773797& n01820546& n01855672& n01978455& n01980166 \\
        \hline
        n01983481& n02009229& n02018207& n02085620& n02086240& n02086910& n02087046& n02089867& n02089973& n02090622 \\
        \hline
        n02091831& n02093428& n02099849& n02100583& n02104029& n02105505& n02106550& n02107142& n02108089& n02109047 \\
        \hline
        n02113799& n02113978& n02114855& n02116738& n02119022& n02123045& n02138441& n02172182& n02231487& n02259212 \\
        \hline
        n02326432& n02396427& n02483362& n02488291& n02701002& n02788148& n02804414& n02859443& n02869837& n02877765 \\
        \hline
        n02974003& n03017168& n03032252& n03062245& n03085013& n03259280& n03379051& n03424325& n03492542& n03494278 \\
        \hline
        n03530642& n03584829& n03594734& n03637318& n03642806& n03764736& n03775546& n03777754& n03785016& n03787032 \\
        \hline
        n03794056& n03837869& n03891251& n03903868& n03930630& n03947888& n04026417& n04067472& n04099969& n04111531 \\
        \hline
        n04127249& n04136333& n04229816& n04238763& n04336792& n04418357& n04429376& n04435653& n04485082& n04493381 \\
        \hline
        n04517823& n04589890& n04592741& n07714571& n07715103& n07753275& n07831146& n07836838& n13037406& n13040303 \\
        \noalign{\hrule height 1pt}
	\end{tabular}
    \vspace{0.2cm}
	\caption{All 100 labels of the ImageNet100 dataset. We take a subset of ImageNet datasets, and list the 100 labels here.}
\end{table*}

\begin{figure*}[t]
\centering
\includegraphics[width=1.0\linewidth, height=5.7cm]{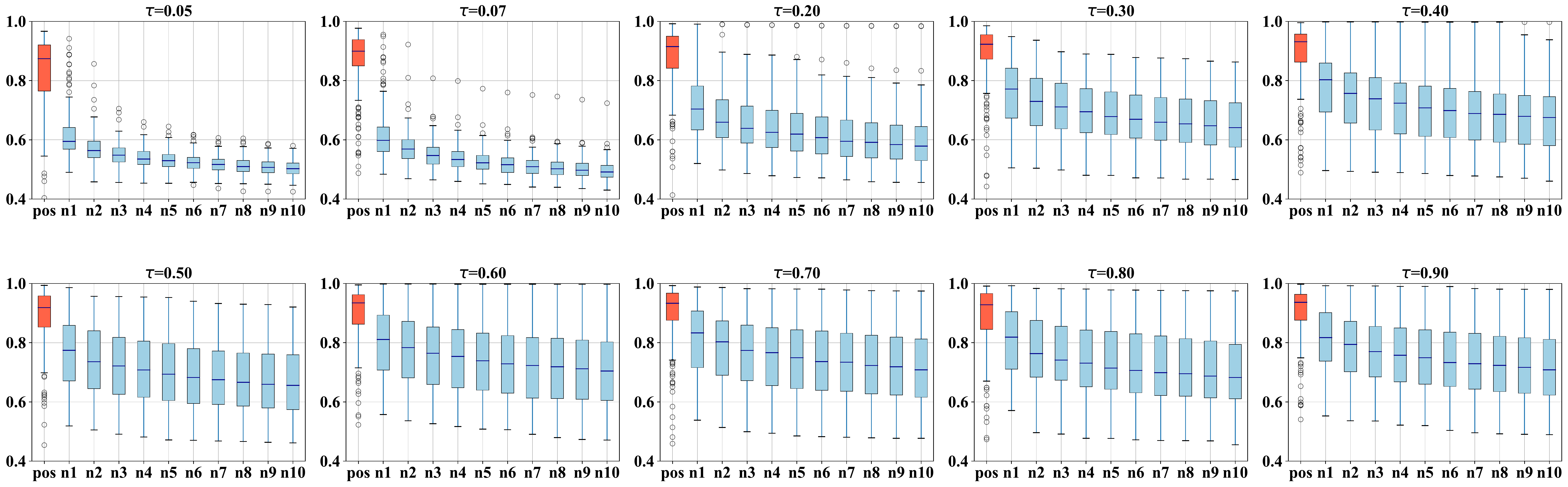}
\caption{We display the similarity distribution of positive samples marked as 'pos' and the distributions of the top-10 nearest negative samples marked as 'ni' for the i-th nearest neighbour. All models are trained with the ordinary contrastive loss on CIFAR10. }
\label{fig: exp1}
\end{figure*}

\begin{figure*}[t]
\centering
\includegraphics[width=1.0\linewidth, height=5.7cm]{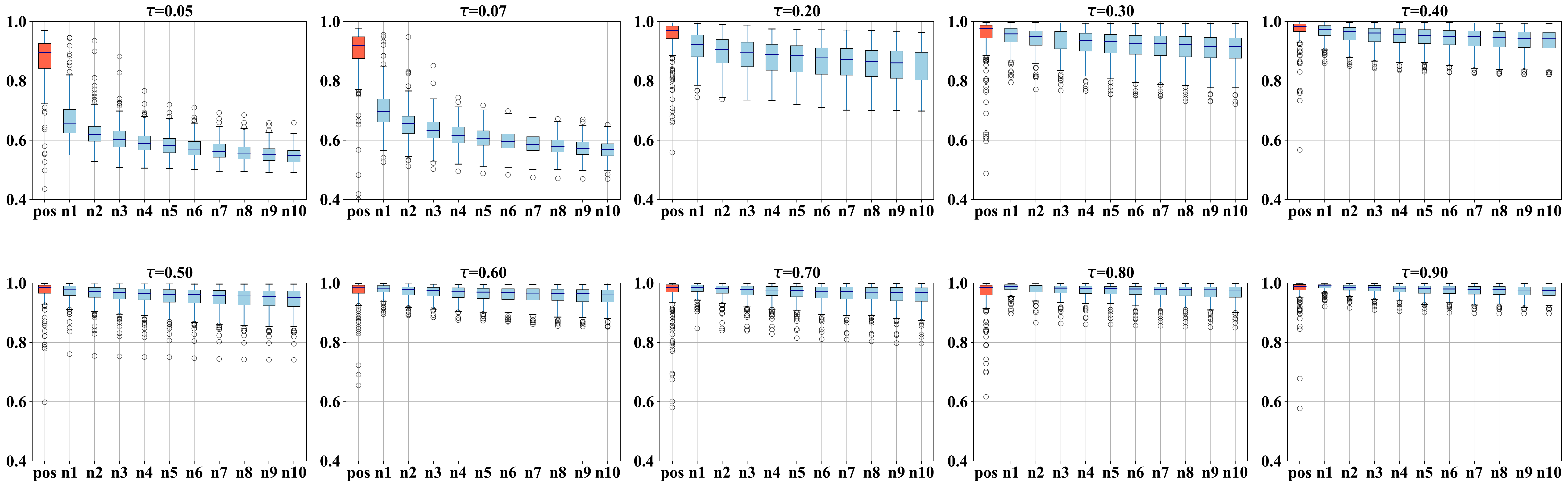}
\caption{We display the similarity distribution of positive samples marked as 'pos' and the distributions of the top-10 nearest negative samples marked as 'ni' for the i-th nearest neighbour. All models are trained with the ordinary contrastive loss on SVHN. }
\label{fig: exp2}
\end{figure*}

\begin{figure*}[t]
\centering
\includegraphics[width=1.0\linewidth, height=5.7cm]{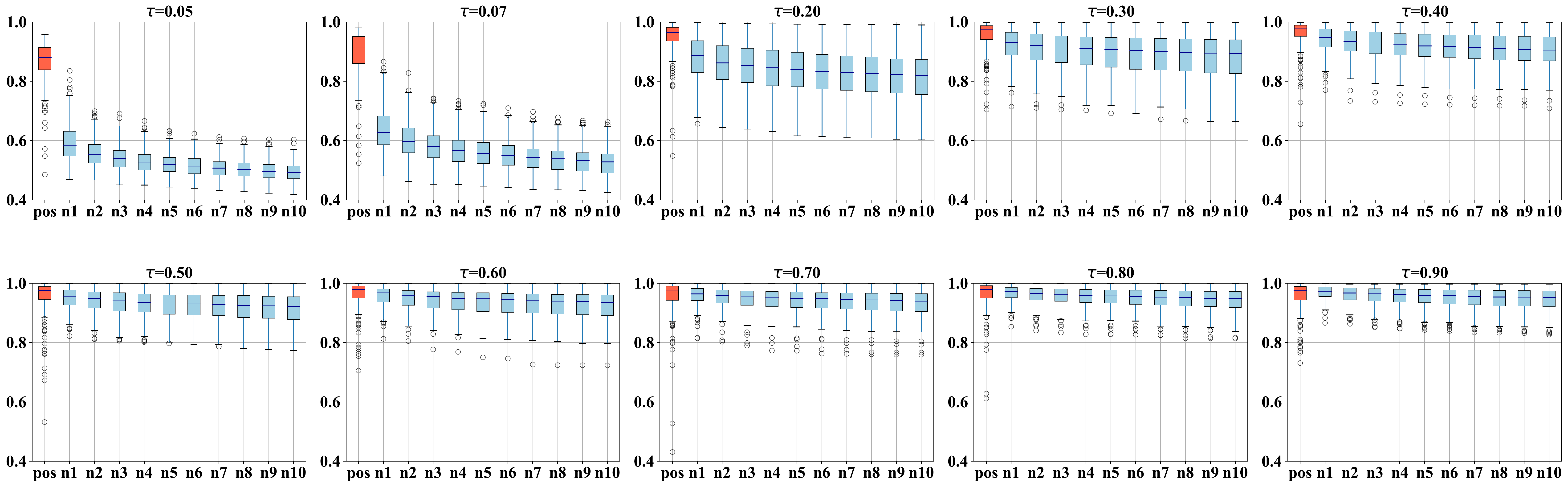}
\caption{We display the similarity distribution of positive samples marked as 'pos' and the distributions of the top-10 nearest negative samples marked as 'ni' for the i-th nearest neighbour. All models are trained with the ordinary contrastive loss on ImageNet100. }
\label{fig: exp3}
\end{figure*}

\begin{figure*}[t]
\centering
\includegraphics[width=1.0\linewidth, height=5.7cm]{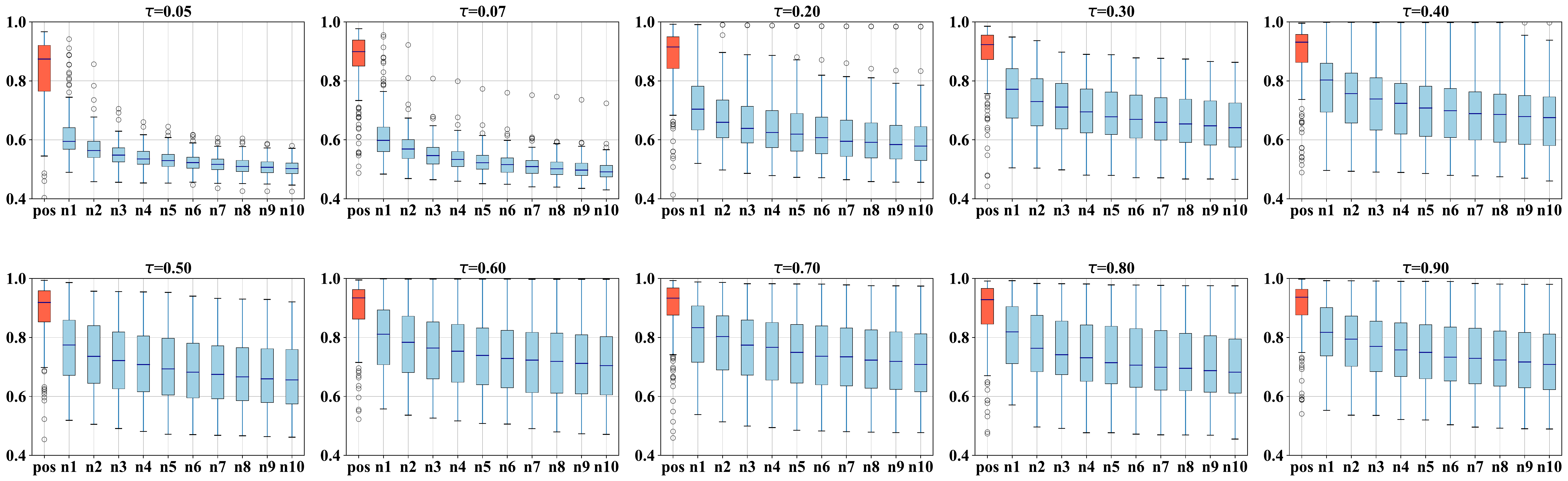}
\caption{We display the similarity distribution of positive samples marked as 'pos' and the distributions of the top-10 nearest negative samples marked as 'ni' for the i-th nearest neighbour. All models are trained with the \textbf{hard} contrastive loss on CIFAR10.}
\label{fig: exp4}
\end{figure*}

\begin{figure*}[t]
\centering
\includegraphics[width=1.0\linewidth, height=5.7cm]{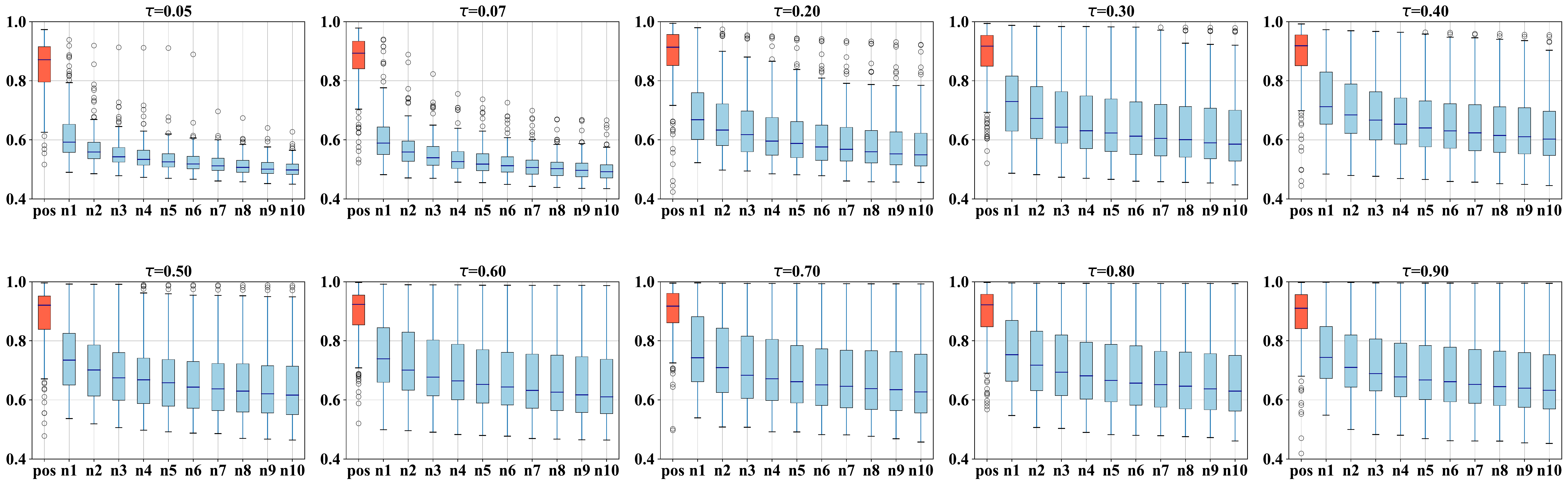}
\caption{We display the similarity distribution of positive samples marked as 'pos' and the distributions of the top-10 nearest negative samples marked as 'ni' for the i-th nearest neighbour. All models are trained with the \textbf{hard} contrastive loss on CIFAR100. }
\label{fig: exp5}
\end{figure*}

\begin{figure*}[t]
\centering
\includegraphics[width=1.0\linewidth, height=5.7cm]{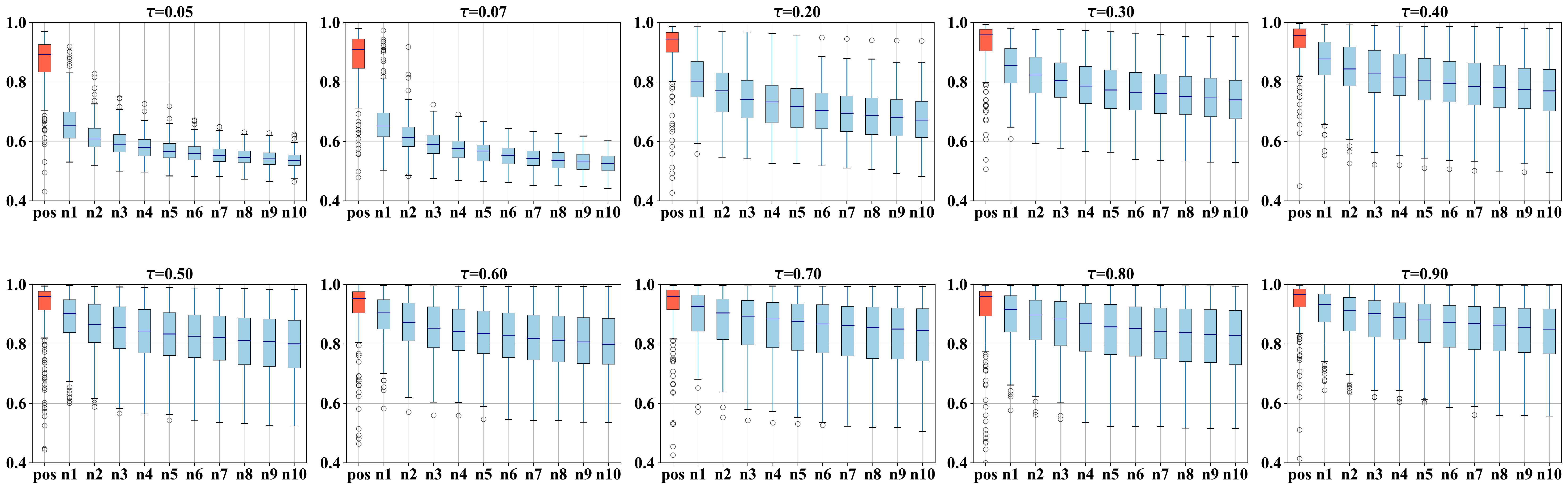}
\caption{We display the similarity distribution of positive samples marked as 'pos' and the distributions of the top-10 nearest negative samples marked as 'ni' for the i-th nearest neighbour. All models are trained with the \textbf{hard} contrastive loss on SVHN. }
\label{fig: exp6}
\end{figure*}

\begin{figure*}[t]
\centering
\includegraphics[width=1.0\linewidth, height=5.7cm]{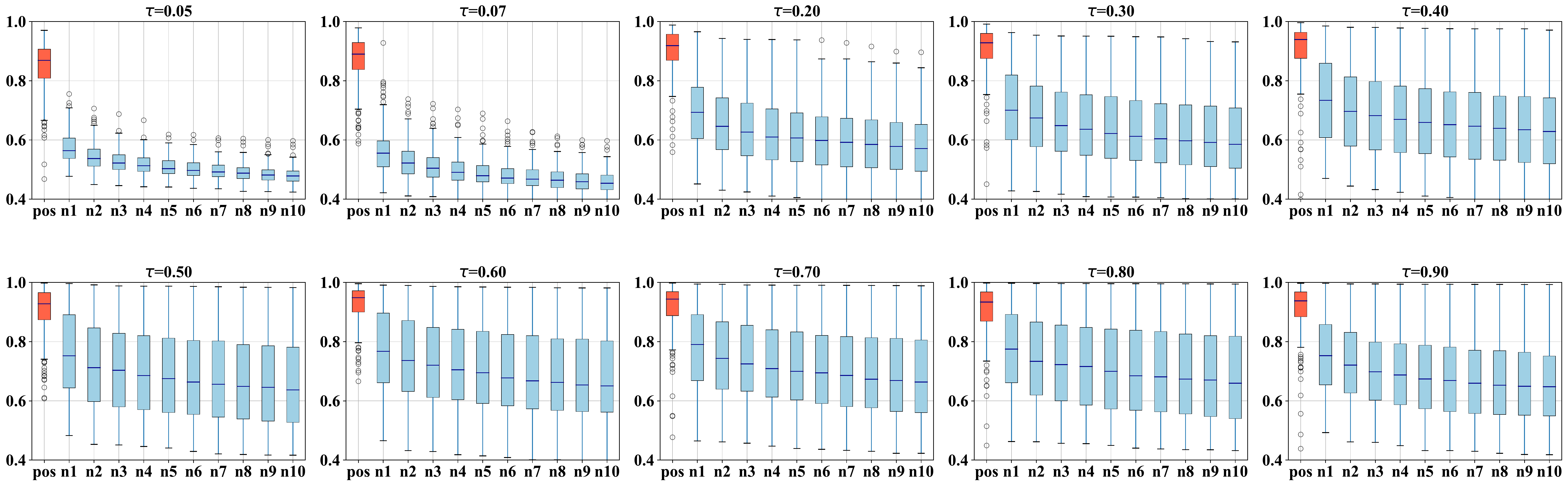}
\caption{We display the similarity distribution of positive samples marked as 'pos' and the distributions of the top-10 nearest negative samples marked as 'ni' for the i-th nearest neighbour. All models are trained with the \textbf{hard} contrastive loss on ImageNet100. }
\label{fig: exp7}
\end{figure*}

{\small
\bibliographystyle{ieee_fullname}
\bibliography{egbib}
}